\documentclass[conference]{IEEEtran}
\IEEEoverridecommandlockouts

\usepackage{cite}
\usepackage{amsmath,amssymb,amsfonts}
\usepackage{algorithmic}
\usepackage{graphicx}
\usepackage{textcomp}
\usepackage{xcolor}

\usepackage{booktabs}
\usepackage{multirow}
\usepackage{subcaption}
\usepackage[letterpaper,margin=0.75in]{geometry}

\def\BibTeX{{\rm B\kern-.05em{\sc i\kern-.025em b}\kern-.08em
    T\kern-.1667em\lower.7ex\hbox{E}\kern-.125emX}}
\begin{document}

\title{Towards Reliable and Robust LLM Planning: \\ A Symbolic Feedback-Driven Iterative Self-Refinement Framework\\
\thanks{* Corresponding author at: Institute of Automation, Chinese Academy of Sciences, Beijing, China.}
}
\author{
    \IEEEauthorblockN{Jiajing Zhang$^{1}$, Jiamei Jiang$^{1,2}$, Chenyang Zhang$^{1}$, Feifei Mo$^{1,3}$, Linjing Li$^{1*}$, Daniel Zeng$^1$}
    \IEEEauthorblockA{$^1$ State Key Laboratory of Multimodal Artificial Intelligence Systems, \\ Institute of Automation,  Chinese Academy of Sciences, Beijing, China}
    \IEEEauthorblockA{$^2$ School of Artificial Intelligence, University of Chinese Academy of Sciences, Beijing, China}
    \IEEEauthorblockA{$^3$ School of Industry-education Integration, University of Chinese Academy of Sciences, Beijing, China}
    \IEEEauthorblockA{jiajing.zhang@ia.ac.cn, jiangjiamei2024@ia.ac.cn, chenyang.zhang@ia.ac.cn, mofeifei2025@ia.ac.cn, \\ linjing.li@ia.ac.cn, dajun.zeng@ia.ac.cn}}
\maketitle

\begin{abstract}
Large language models (LLMs) have attracted widespread attention from academia and industry, yet their deployment raises critical security concerns regarding robustness and reliability. Planning, a core component of intelligent behavior, remains challenging for LLMs, which often produce infeasible or incorrect solutions in long-horizon decision-making tasks due to inherent complexity. In this paper, we propose a symbolic feedback-driven iterative self-refinement framework to enhance the robustness and reliability of LLMs in long-horizon planning. Specifically, a natural language prompting mechanism is introduced to map logical symbols into natural language descriptions, enabling LLMs to better capture task constraints and semantics. We further design a symbolic verifier that identifies errors and converts them into corrective instructions interpretable by the LLM, thereby guiding self-refinement. In addition, we leverage a plan recognizer to infer goal reachability, facilitating more effective guidance toward desired goals. Empirical results demonstrate that the proposed framework consistently improves both feasibility and correctness in long-horizon planning tasks. This highlights its effectiveness in enhancing the reliability of LLM-based planning and potential to enable more trustworthy AI systems. 

\end{abstract}

\section{INTRODUCTION}

Large language models (LLMs) have significantly advanced natural language understanding and generation, attracting widespread attention from academia and industry, while also raising critical security concerns regarding robustness, reliability, and susceptibility to hallucinations \cite{wei2025plangenllms,betley2026training}. Planning, as a fundamental component of intelligent behavior, is essential for ensuring reliable and verifiable AI systems. However, empirical evidence indicates that LLMs remain challenged by planning tasks, particularly those requiring long-horizon reasoning \cite{duan2024gtbench}. These limitations stem from the intrinsic complexity of long-horizon planning \cite{BYLANDER1994165}, which involves constructing sequences of interdependent actions over extended horizons to achieve control objectives \cite{zhou2024isr}. Unlike single-step reasoning, long-horizon planning must account for the delayed and compounding effects of early decisions \cite{guptaComplexityBlocksworldPlanning1992}. Such characteristics impose significant challenges in modeling long-range temporal dependencies and exacerbate combinatorial complexity. Consequently, effective planning in this setting requires carefully balancing efficiency, optimality, and adaptability \cite{hartmann2022long}.

Despite their strong language capabilities, LLMs often suffer from inefficiency, susceptibility to hallucinations, and a lack of guarantees for feasibility and correctness in planning \cite{valmeekam2024llmscantplanlrms,kambhampati2024can}. These limitations raise critical security concerns, as their planning strategies may violate constraints, leading to contradictions and conflicts between actions and states \cite{wang2026reasoning}. Moreover, LLM-based planners tend to select locally plausible actions without considering long-term consequences, causing errors to accumulate and often resulting in systematic failures in long-horizon tasks \cite{XU2025101370}. Consequently, developing advanced methods to enhance secure and reliable long-horizon planning in LLM-based agents remains a critical challenge for building trustworthy AI systems. Classical symbolic planners provide a well-established paradigm for planning by representing problems using explicit symbols, logic, and rules, enabling human-interpretable and high-level reasoning \cite{blum1997fast}. However, these approaches rely heavily on expert-crafted domain knowledge and are restricted to structured symbolic inputs, limiting their ability to process natural language instructions or implicit constraints \cite{yuSurveyNeuralsymbolicLearning2023,bonetPlanningHeuristicSearch2001a}. As a result, they exhibit limited usability, which constrains their applicability in diverse real-world scenarios \cite{jendrikFastDownwardScorpion}.

In summary, LLMs excel in language understanding but lack reliable planning capabilities, while symbolic planners provide rigorous and interpretable reasoning yet suffer from limited flexibility. This complementarity motivates the development of neural-symbolic approaches that integrate LLMs with symbolic methods to leverage their respective strengths \cite{pnas.2516995122}. In such frameworks, symbolic components provide precise and interpretable verification signals to guide planning, improving feasibility and correctness, while LLMs enable flexible language capabilities, reducing reliance on manually specified symbolic representations \cite{zhang2025closed,10.24963/ijcai.2025/1195}.

\begin{figure*}[!tbp]
\centering
\includegraphics[width=1.0\linewidth]{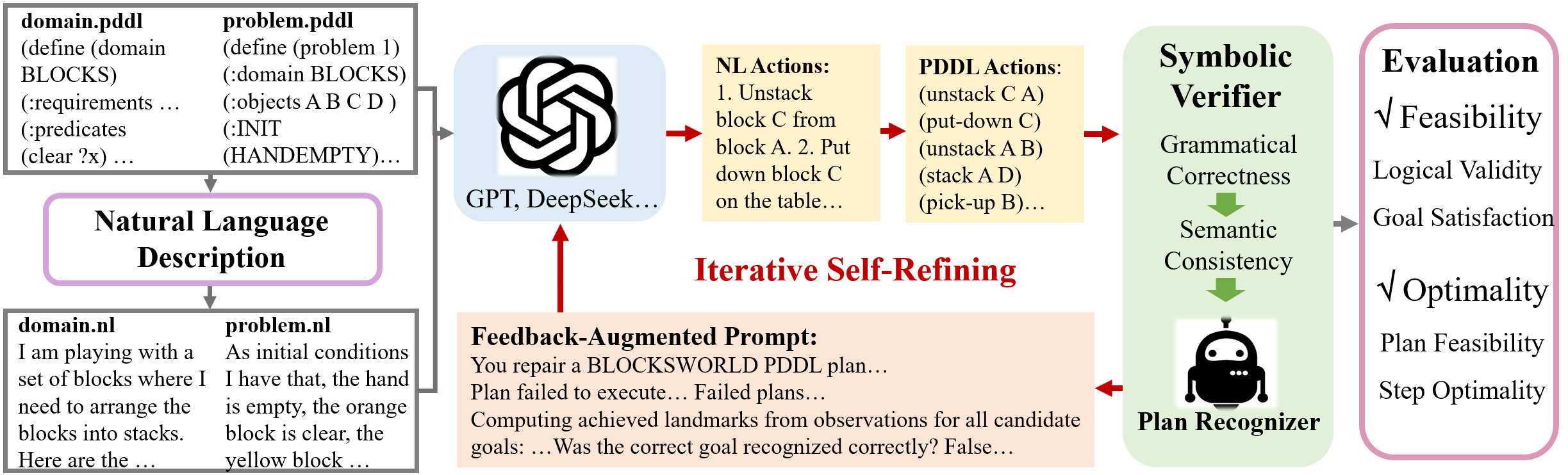}
\caption{The overview of the proposed symbolic feedback-driven iterative self-refinement framework.}
\label{fig1}
\end{figure*}

This study proposes a symbolic feedback-driven iterative self-refinement framework to enhance the robustness and reliability of LLMs in long-horizon planning. It leverages a neural-symbolic feedback mechanism to integrate structured knowledge into the planning process, enabling accurate error detection and correction. Specifically, a symbolic verifier interacts with the LLMs to evaluate generated plans and ensure goal satisfaction, thereby improving feasibility and correctness. When errors such as action conflicts or constraint violations are detected, the verifier provides structured feedback to guide corrective reasoning, mitigating hallucinations and enhancing planning reliability. In addition, the framework incorporates a plan recognizer to infer goal reachability, facilitating effective guidance toward desired goals. Our empirical results validate that the proposed framework significantly improves the performance of LLMs in long-horizon planning tasks. Moreover, the diverse feedback strategies embedded in our framework prove effective, providing flexibility and generality across different LLM-based planners and planning scenarios. The results show that the framework can consistently enhance both the robustness and reliability of LLM-generated plans, which are crucial for enabling more reliable and trustworthy AI systems.

Our main contributions are summarized as follows:
\begin{itemize}
\item We propose a symbolic feedback-driven iterative self-refinement framework that enhances the robustness and reliability of LLMs.
\item We design a symbolic verifier that translates errors into corrective instructions interpretable by LLMs, providing precise guidance for self-refinement.
\item We incorporate a plan recognizer to infer goal reachability and facilitate symbolic knowledge transfer, steering the planning process toward the intended goal.
\item Extensive experiments demonstrate that the proposed framework consistently improves long-horizon planning capabilities of LLMs.
\end{itemize}


\section{BACKGROUND}
\subsection{Definition of Planning} 
Planning refers to the process of generating a sequence of actions that enables an agent to achieve the desired goals\cite{ghallab2004automated}.
Formally, a planning problem can be defined as a triple $ P=(\Xi,\mathcal{I},G_{tru}) $, where denotes the domain specification $ \Xi $ , consisting of a set of state variables $ \mathcal{V} $ and a set of actions $ A $. $ \mathcal{I} $ represents the initial state, while $ G_{tru} $ denotes the planning goal \cite{allmendinger2017planning}. An action $a\in\mathcal{A}$ is denoted by a triple $ \left({name}_a,{pre}_a,{eff}_a\right)$, where ${name}_a$ represents the description of $a$; ${pre}_a$ and ${eff}_a$ called its preconditions and effects. A state $s \in \mathcal{V} $ is a finite set of positive facts $f$ that follows the world assumption so that if $f\in s$, then $f$ is true in $s$. An action $a$ is applicable in a state $s$ if ${pre}_a\in s$. Applying an action $a$ in a state $s$ leads to a new state $s^\prime$ with the same variable as in $s$, except for those variables defined in ${eff}_a$. A planner $\Pi$ solves a confronted planning problem $P$ by crafting an appropriate plan $\pi=\Pi(P)$ usually represented as a sequence of actions $\pi=\ \left(a_0, a_1, \ldots, a_n\right) $, which transforms the initial state $ \mathcal{I} $ into the goal state $ G_{tru} $ by sequentially applying those actions in $ \pi $. 

\subsection{Planning Domain Definition Language} 

Planning is typically formulated using the Planning Domain Definition Language (PDDL), which provides a unified standard for representing planning problems \cite{haslum2019introduction}. PDDL represents states and actions based on predicates\cite{aeronautiques1998pddl}. An n-ary predicate $ p $ is applied to a sequence of terms $\left(t_1,\ldots,t_n\right)$, where each term is either a constant or a variable, representing an object in the domain. Grounded predicates, referred to as facts, correspond to predicates instantiated with specific objects. The structure of PDDL is generally divided into two components: domain definition and problem definition. The domain definition captures the shared characteristics of a class of planning problems, including type specifications, predicate definitions, and action schemas. In contrast, the problem definition specifies a concrete planning instance, including the initial state, goal conditions, and the set of involved objects.

\section{METHODOLOGY}

This section presents the proposed framework illustrated in Fig.~\ref{fig1}. We first introduce the natural language prompting mechanism for planning tasks, followed by the feedback-driven iterative self-refinement framework for LLMs, and finally describe the recognition-based symbolic verifier.

\begin{figure}[!tbp]
	\centering
	\includegraphics[width=0.49\textwidth]{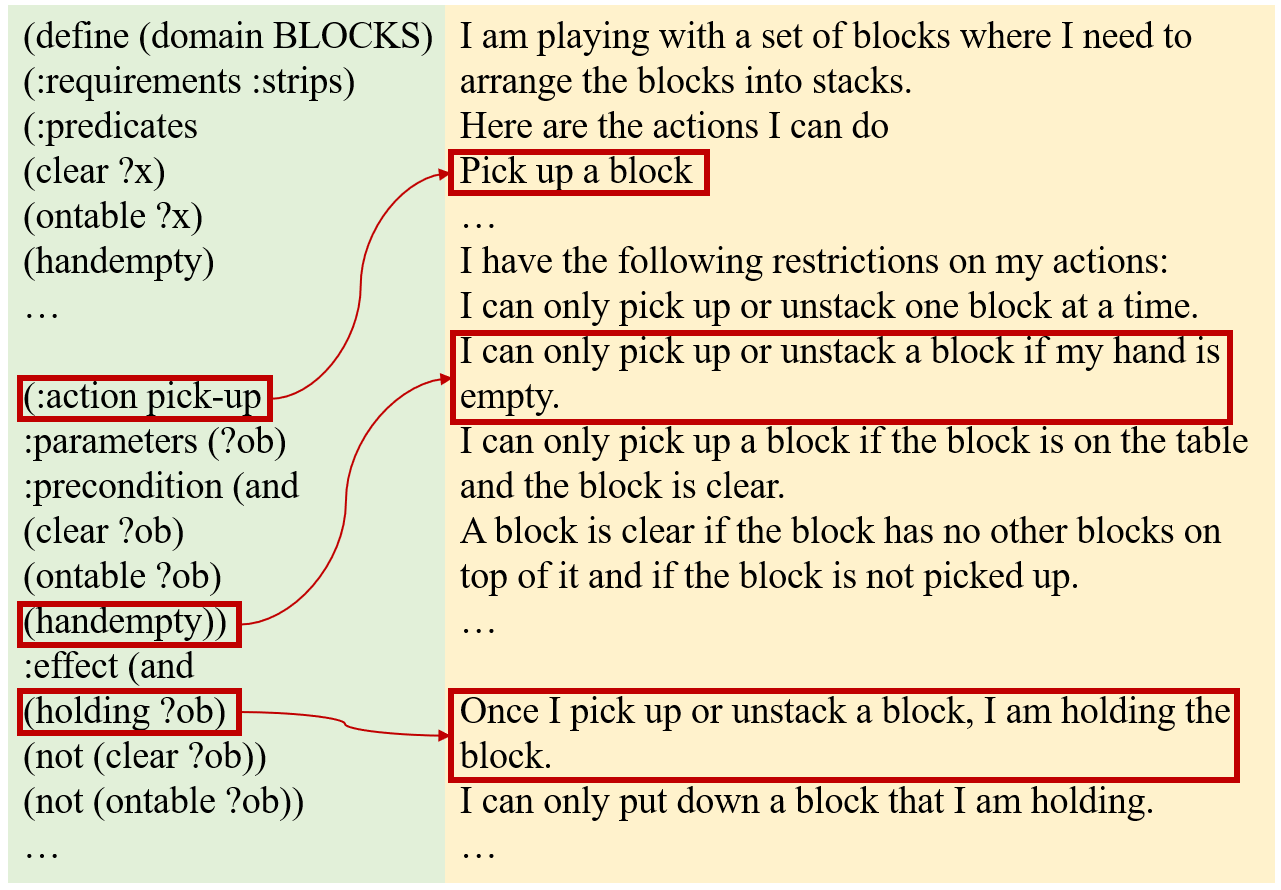}
    \caption{Example of a NL Prompt for Blocksworld Domain.}
	\label{fig2}
	
\end{figure}

\subsection{Natural Language Prompting Mechanism}
Planning tasks often rely on structured symbolic representations in PDDL to enable precise and rigorous reasoning. LLMs excel at natural language (NL) understanding and generation but exhibit limited capability in parsing symbolic languages used in planning\cite{sun2023adaplanner}. As a result, they are prone to issues such as syntactic parsing errors, weak semantic alignment, and misinterpretation of domain rules, which significantly undermine the accuracy and efficiency of planning\cite{stein2025automating}.
This paper proposes a natural language prompting mechanism to establish a mapping between PDDL symbolic language and natural language. It preserves the logical constraints of PDDL while leveraging the semantic intuitiveness of natural language, enabling LLMs to accurately grasp both the constraint boundaries and the semantic content of planning tasks. The construction of such prompts can be formally represented as:

\begin{equation}
prompt = Concat\left( template,  \Xi,\mathcal{I},G_{tru} \right)
\label{eq1} 
\end{equation}
where $template$ refers to a predefined textual structure used to systematically organize problem information into a format that LLMs can process. $ \Xi $ denotes the planning domain definition, while $\mathcal{I}$ and $G_{tru}$ represent the initial state and goal of the planning problem, respectively.

Taking the representative Blocksworld task as an example, we illustrate the above conversion process in detail. As shown in Fig \ref{fig2}, key information such as actions, predicates, and objects is first extracted from the PDDL file and then mapped into NL. Based on the semantics of the Blocksworld domain, the PDDL action $pick-up$ is converted to the NL “Pick up a block”. Its preconditions and effects are similarly translated. For example, the precondition $handempty$ is converted to “I can only pick up or unstack a block if my hand is empty,” clearly specifying the necessary conditions for executing the action. Likewise, the effect $(holding \ ?ob)$ is translated as “Once I pick up or unstack a block, I am holding the block,” indicating the resulting state after the action. The natural language descriptions generated through this mechanism retain the symbolic constraints of the PDDL file while presenting them in a format that LLMs can efficiently understand. This mechanism mitigates the risk of misinterpreting domain rules due to difficulties in parsing symbolic syntax and provides a foundation for subsequent planning by LLMs.

\subsection{Feedback-Driven Iterative Self-Refinement Framework}
As shown in Fig. \ref{fig1}, the PDDL files and natural language description files are fed into the LLMs to generate an initial plans $ \pi $ aimed at achieving the planning goal. However, due to issues such as hallucinations, the initial sequence often contains errors, including action conflicts or goal unreachability. To address this, we propose a feedback-driven self-refinement framework, which constructs an automatic prompt engineering that translates errors identified by the symbolic verifier into corrective instructions understandable by the LLMs, providing precise guidance for the model’s self-optimization process. Specifically, we generate feedback-augmented prompt to optimize the resulting plan of LLMs. The feedback-augmented prompt (Eq. \ref{eq2}) consists of three components: 

\begin{equation}
T^{'} = T_{input} \oplus T_{feedback} \oplus T_{history}
\label{eq2} 
\end{equation}
where the original base prompt $T_{input} = \text{Tokenize}\left( prompt \right)$, contains the tokenized output from the natural language prompting mechanism and includes core information such as the planning task objectives, domain rules, and initial state. The history component, $T_{history}$, records previous planning attempts, while the augmented feedback, $T_{feedback}$ , provides error detection information. 


Based on the feedback-augmented prompts from symbolic verifier, the LLMs corrects erroneous actions in the previous plans $ \pi $ ; and designs new plans $\pi^{'} $ that satisfy the planning goal. This forms an iterative self-optimization loop encompassing prompt refinement, strategy generation, error validation, and feedback updating. The loop terminates under either of the following conditions: (1) the verifier detects no errors and produces a feasible planning strategy; or (2) a preset maximum number of iterations is reached. By constructing a feedback-driven self-refinement framework, this approach establishes a “feedback–refine–guide” closed loop, shifting the model’s self-optimization process from blind trial-and-error to precise, targeted correction.

\begin{figure}[!bp]
	\centering
	\includegraphics[width=0.45\textwidth]{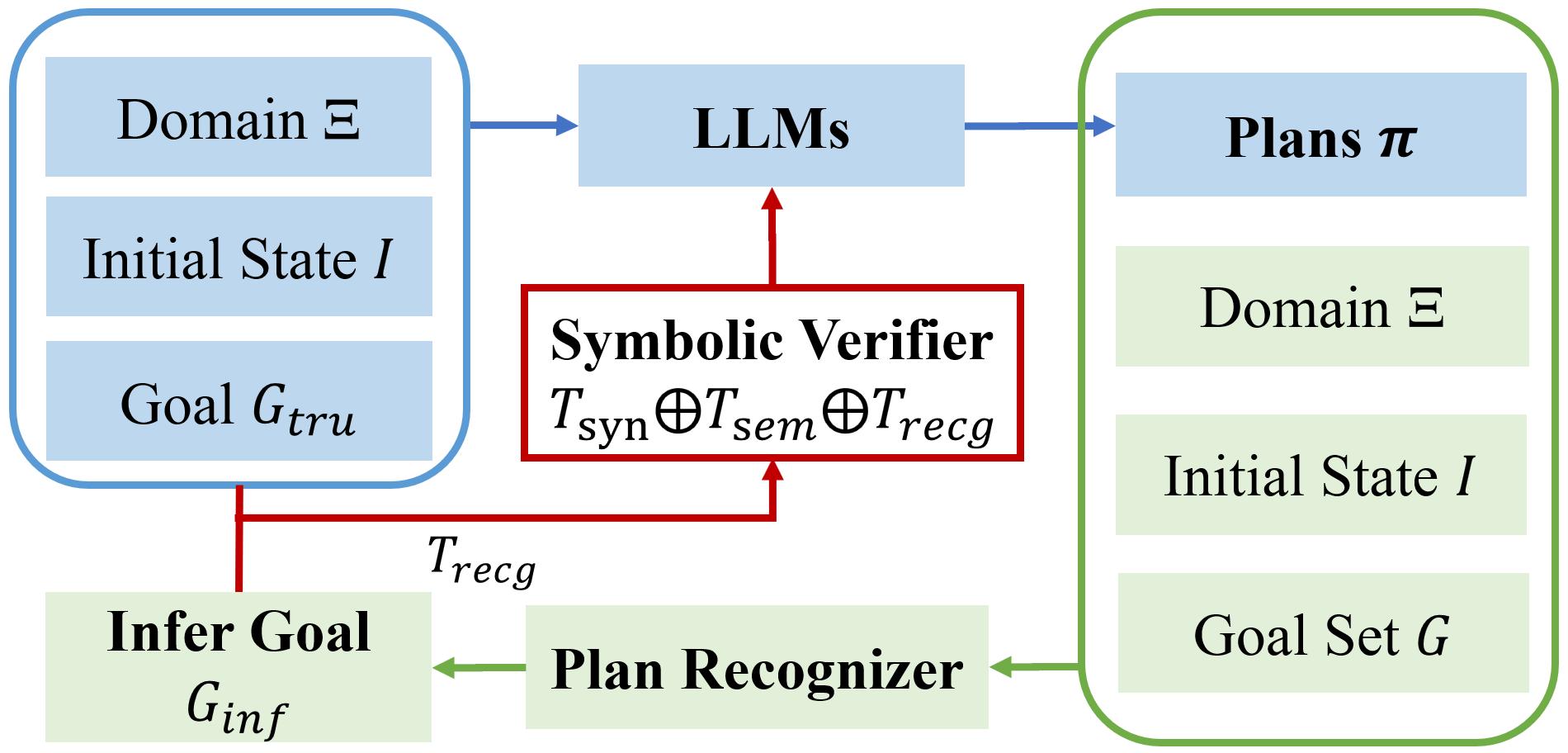}
    \caption{Recognition-based symbolic verifier.}
	\label{fig3}
	
\end{figure}

\subsection{Recognition-based Symbolic verifier}

As central component of the feedback-driven iterative self-refinement framework, the symbolic verifier performs hierarchical and multi-dimensional checks, supplying precise error signals to guide the feedback-enhanced automatic prompting mechanism. First, the generated plans $ \pi $ are checked for syntactic correctness and semantic consistency. For syntax \cite{holler2020hddl}, the focus is on verifying action and object naming conventions as well as parameter matching. For semantics, the VAL validation tool \cite{howey2004val} is employed to determine whether the action sequence can be correctly parsed and executed. After performing syntactic and semantic checks, we employ a symbolic planning recognizer \cite{pereira2017landmark} to infer goal reachability, thereby more effectively guiding the planning process toward the intended goal.

As shown in Fig. \ref{fig3}, the plan recognizer infers the most likely goal\cite{ramirez2009plan} based on the the domain $\Xi$, the initial state $\mathcal{I}$, the set of possible goals $G$, and the a series of observed actions. Our framework involves a symbolic plan recognizer, which first extract useful recognition information $\zeta \left(g_1,\ldots,g_n\right)$  for all candidate goals $ g_i \in G $ from the recognition problem $\left(\mathrm{\Xi},\mathcal{I},G\left(g_1,\ldots,g_n\right),\pi \right)$(Eq. \ref{eq3}). And then the function $ \Phi \left( \cdot \right) $ utilize the recognition calculates the likelihood of each candidate goal information(Eq. \ref{eq4}). Finally, candidate goals are ranked based on their corresponding likelihoods, represented by $ (\psi \left(g_1,\ldots,g_n\right) $, and the goal with the highest likelihood is selected as the output, $ G_{inf} $ (Eq. \ref{eq5}).

\begin{equation}
\left(\mathrm{\Xi},\mathcal{I},G\left(g_1,\ldots,g_n\right),\pi \right) \rightarrow \zeta \left(g_1,\ldots,g_n\right)
\label{eq3} 
\end{equation}
\begin{equation}
\Phi \left( \zeta \left(g_1,\ldots,g_n\right) \right) \rightarrow \psi \left(g_1,\ldots,g_n\right)
\label{eq4} 
\end{equation}
\begin{equation}
max\left(\psi\left(g_1,\ldots,g_n\right)\right) \rightarrow G_{inf} \rightarrow T_{recg}
\label{eq5} 
\end{equation}
\begin{equation}
T_{feedback} = T_{syn} \oplus T_{sem} \oplus T_{recg}
\label{eq6} 
\end{equation}
 
By comparing the inferred goal $G_{inf}$ produced by the recognizer with the true goal $G_{tru}$ of the original planning problem, we can assess whether the generated action sequence effectively achieves the intended goal. If the inferred goal matches the true goal, the action sequence is considered goal-consistent; otherwise, a goal deviation error is identified. In such cases, the discrepancy is recorded and transformed into augmented feedback $T_{recg}$. The final output of the symbolic verifier is a multi-dimensional feedback $T_{feedback}$ (Eq. \ref{eq6}), which consists of three components: syntactic correctness feedback $T_{syn}$, semantic consistency feedback $T_{sem}$, and goal reachability feedback $T_{recg}$ based on planning recognition. This feedback is then fed into the feedback-enhanced prompting, providing precise error information for dynamic prompt optimization and the self-refinement of LLMs.

\section{EXPERIMENTS}
In this section, we conduct comparative experiments to evaluate the performance. We also present a detailed analysis to demonstrate the effectiveness of our framework. The hardware consisted of a desktop machine running Ubuntu 24.04, with 1.0 TB of RAM and AMD EPYC 9684X processors.

\subsection{Experimental Settings}

To comprehensively evaluate the performance of the proposed framework, we adopt a multi-dimensional evaluation based on feasibility and optimality. A plan is feasible if every action satisfies its preconditions and the sequence transforms the initial state into the goal state; it is optimal if it is feasible and has minimal length. We evaluate performance using planning coverage (success rate), defined as the proportion of problems solved relative to the total number of problems. 
Experiments are conducted on the PlanBench benchmark\cite{10.5555/3666122.3667815}, which comprises 1,200 tasks and is widely used to evaluate the planning capabilities of LLMs.
In addition, we construct a plan recognizer based on a landmark-based method \cite{pereira2017landmark}, which serves as the symbolic verifier within our framework. The verifier terminates if the LLM-generated plans are feasible/optimal solutions, or after completing 5 maximum refinement rounds.

\subsection{Comparison of Performance}
To validate the superior planning capabilities of our framework, we conducted comparative experiments with three mainstream LLMs: GPT-4o, Claude-3-5, and DeepSeek-R1, using standard inference parameters (temperature = 0.1) for consistency. Two evaluation modes are considered: LLM-direct, in which the model generates planning solutions directly, and LLM-feedback, which integrates our symbolic feedback-driven iterative self-refinement framework. Experiments are conducted in two representative domains: Blocksworld, a classical planning scenario with clear task logic, and Mystery, a challenging variant in which predicates and states (e.g., “handempty”, “pick-up”) are replaced with arbitrary symbols, removing explicit semantic information.
\begin{table}[!bp]
	\caption{RESULTS OF DIFFERENT LLMS}

	\centering
	\begin{tabular}{cccccc}
			\hline
			\textbf{LLMs} & \multicolumn{3}{c}{\textbf{Method}} & \textbf{Blocksworld} & \textbf{Mystery}  \\
			\hline
			\multirow{2}{*}{GPT-4o}& \multicolumn{3}{l}{LLM-direct} & 35.5\% & 0.0\% \\
			& \multicolumn{3}{l}{LLM-fb}  & 69.5\% ($\uparrow$34.0\%)  & 3.8\% ($\uparrow$3.8\%) \\
			\hline

			\multirow{2}{*}{Claude-3-5}& \multicolumn{3}{l}{LLM-direct} & 54.8\% & 0.0\% \\
			& \multicolumn{3}{l}{LLM-fb}  & 83.5\% ($\uparrow$28.7\%)  & 4.0\% ($\uparrow$4.0\%) \\
			\hline
			
			\multirow{2}{*}{DeepSeek-R1}& \multicolumn{3}{l}{LLM-direct} & 99.1\% & 43.3\%  \\
			& \multicolumn{3}{l}{LLM-fb}  & 100.0\%($\uparrow$0.9\%) & 65.5\% ($\uparrow$22.2\%) \\
			\hline

		\end{tabular}
		\label{tab1}
\end{table}

As shown in Table \ref{tab1}, the substantial performance differences between the two modes underscore the effectiveness of our framework. In the Blocksworld domain, all models exhibited improvements in plan coverage after incorporating feedback (“LLM-fb” in Table \ref{tab1}). GPT-4o achieved the largest gain, with coverage increasing by 34.0\%, while Claude-3-5 improved by 28.7\%. Following the framework’s optimization, DeepSeek-R1 reached perfect planning performance, achieving 100\% coverage. In the Mystery domain, the results highlight both the planning limitations of LLMs under symbol–semantic confusion and the efficacy of our framework. In LLM-direct mode, GPT-4o and Claude-3-5 achieved 0\% coverage, indicating that without structured feedback, they cannot map arbitrary symbols to predicate semantics. With feedback, both models overcame this limitation, achieving coverage increases of 3.8\% and 4.0\%, respectively, as the structured feedback provides critical cues for error localization and semantic association. DeepSeek-R1’s coverage increased by 22.2\%, demonstrating that the framework not only improves the reliability of planning but also enhances robustness in semantic extraction tasks.

\subsection{Ablation Studies}
To illustrate the impact of the different components of our framework, a series of ablation experiments were conducted in Blockworld domain, with the same setup applied in subsequent experiments. We considered the following ablations:  (1) only the LLM was used to generate plan(``LLM-direct"); (2) the natural language prompt was added (``LLM-NL"); (3) both the natural language prompt and augmented feedback were used (``LLM-NL-AF" ). 
As shown in Table \ref{tab2}, incorporating natural language prompts increased feasibility coverage by a maximum of 15.2\% and optimality coverage by a maximum of 6.5\% compared to the baseline LLM using only the original PDDL files. This demonstrates that mapping PDDL statements into natural language phrases lowers the model’s barrier to understanding symbolic rules, enhancing task comprehension and overall performance. Building on natural language prompting, the feedback-enhanced mechanism—which combines a planning-aware symbolic verifier with dynamic prompt optimization—further improved performance, raising feasibility coverage by a maximum of 14.5\% and optimality coverage by a maximum of 9.0\%. This demonstrates that precise error signals from the augmented feedback guide the LLMs to correct root causes, enhancing planning reliability and strengthening convergence toward optimal solutions.

\begin{table}[!htbp]
	\caption{ABLATION STUDY OF COMPONENTS}

	\centering
	\begin{tabular}{cccccc}
			\hline
			\textbf{LLMs} & \multicolumn{3}{c}{\textbf{Method}} & \textbf{Feasibility } & \textbf{Optimality}  \\
			\hline
			\multirow{3}{*}{GPT-4o}& \multicolumn{3}{l}{LLM-direct} & 53.3\% & 41.0\% \\
			& \multicolumn{3}{l}{LLM-NL} & 55.0\% ($\uparrow$1.7\%) & 42.0\% ($\uparrow$1.0\%)  \\
			& \multicolumn{3}{l}{LLM-NL-AF}  & 69.5\% ($\uparrow$14.5\%)  & 49.8\% ($\uparrow$7.8\%) \\
			\hline

			\multirow{3}{*}{Claude-3-5}& \multicolumn{3}{l}{LLM-direct} & 65.0\% & 47.0\% \\
			& \multicolumn{3}{l}{LLM-NL} & 80.2\% ($\uparrow$15.2\%) & 53.5\% ($\uparrow$6.5\%)  \\
			& \multicolumn{3}{l}{LLM-NL-AF}  & 94.7\% ($\uparrow$14.5\%)  & 62.5\% ($\uparrow$9.0\%) \\
			\hline
			
			\multirow{3}{*}{DeepSeek-R1}& \multicolumn{3}{l}{LLM-direct} & 94.0\% & 91.3\% \\
			& \multicolumn{3}{l}{LLM-NL} & 94.0\% ($\uparrow$0.0\%) & 91.3\% ($\uparrow$0.0\%)  \\
			& \multicolumn{3}{l}{LLM-NL-AF}  & 100.0\% ($\uparrow$6\%)  & 97.2\% ($\uparrow$5.9\%) \\
			\hline

		\end{tabular}
		\label{tab2}
\end{table}

\begin{figure}[!htbp]
	\centering
	\includegraphics[width=0.49\textwidth]{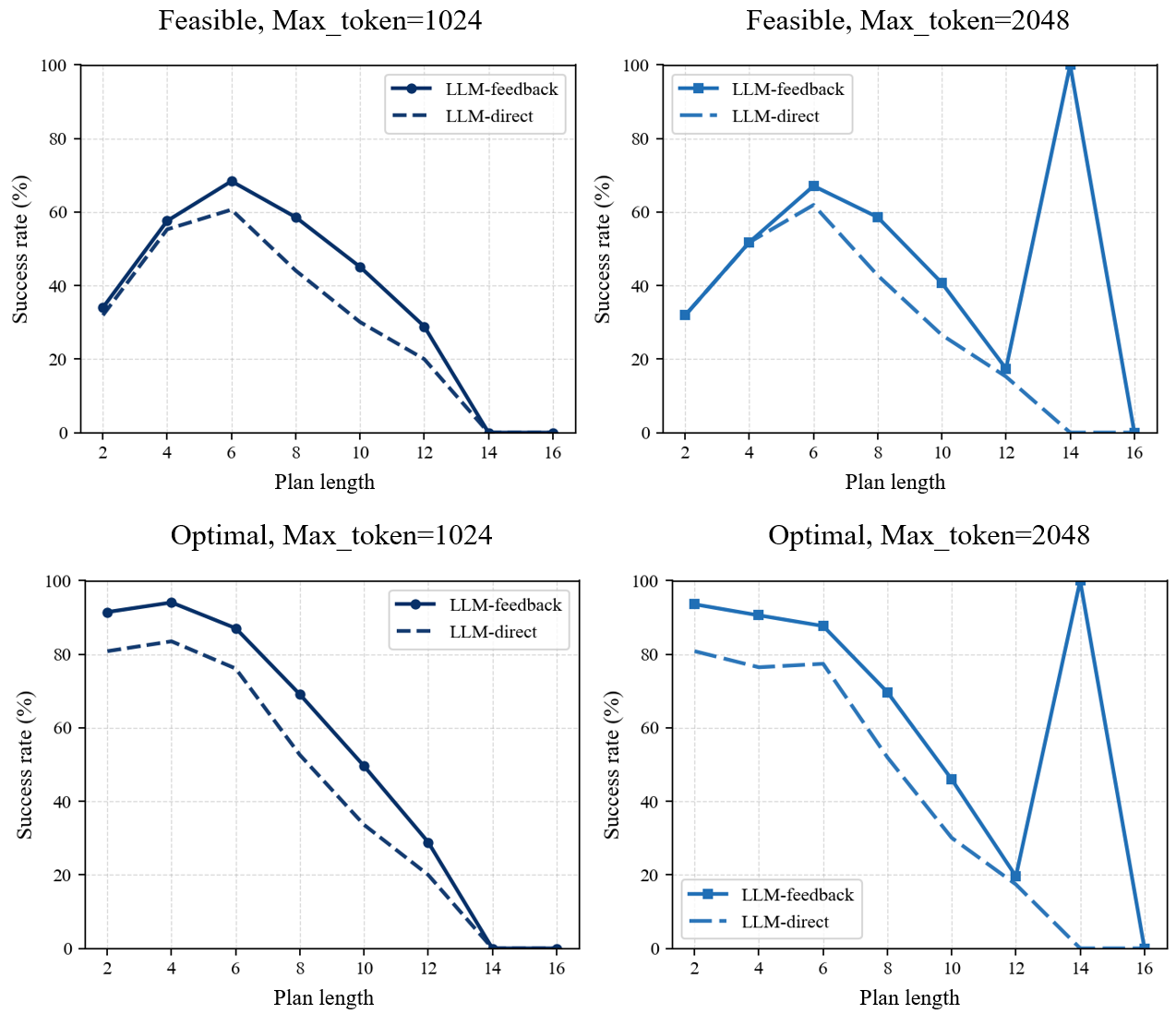}
	\caption{The influence of planning length. }
	\label{fig5}
	
\end{figure}

\subsection{Analysis of Planning Length}
To evaluate the proposed framework across varying plan lengths, we conduct experiments spanning lengths from 2 to 16, analyzing its effectiveness in improving planning accuracy and mitigating long-horizon planning limitations in LLMs. As shown in Fig.~\ref{fig5}, feasible and optimal successful rate declines with increasing plan length for all models. However, the feedback-enhanced model consistently outperforms the baseline across lengths 2–14, with a widening gap as difficulty increases (feasible peaking at 16.5\% at length 6 and optimal peaking at 15.0\% at length 10). This gain stems from correcting errors via real-time symbolic verifier. The results show that all models perform poorly on more challenging instances with plan lengths 14 and 16. To examine whether this is due to token limitations, we increase the maximum token length from 1024 to 2048. As shown in Fig. \ref{fig5}, the baseline models show little improvement, whereas the feedback-enhanced models achieve substantial gains. With sufficient token capacity, feedback effectively guides the LLM in resolving long-horizon dependencies, while the baseline, lacking corrective signals, remains bottlenecked. These results confirm that dynamic error correction via the symbolic verifier significantly improves planning accuracy and robustness in long-horizon planning.

\begin{figure}[!bp]
	\centering
	\includegraphics[width=0.4\textwidth]{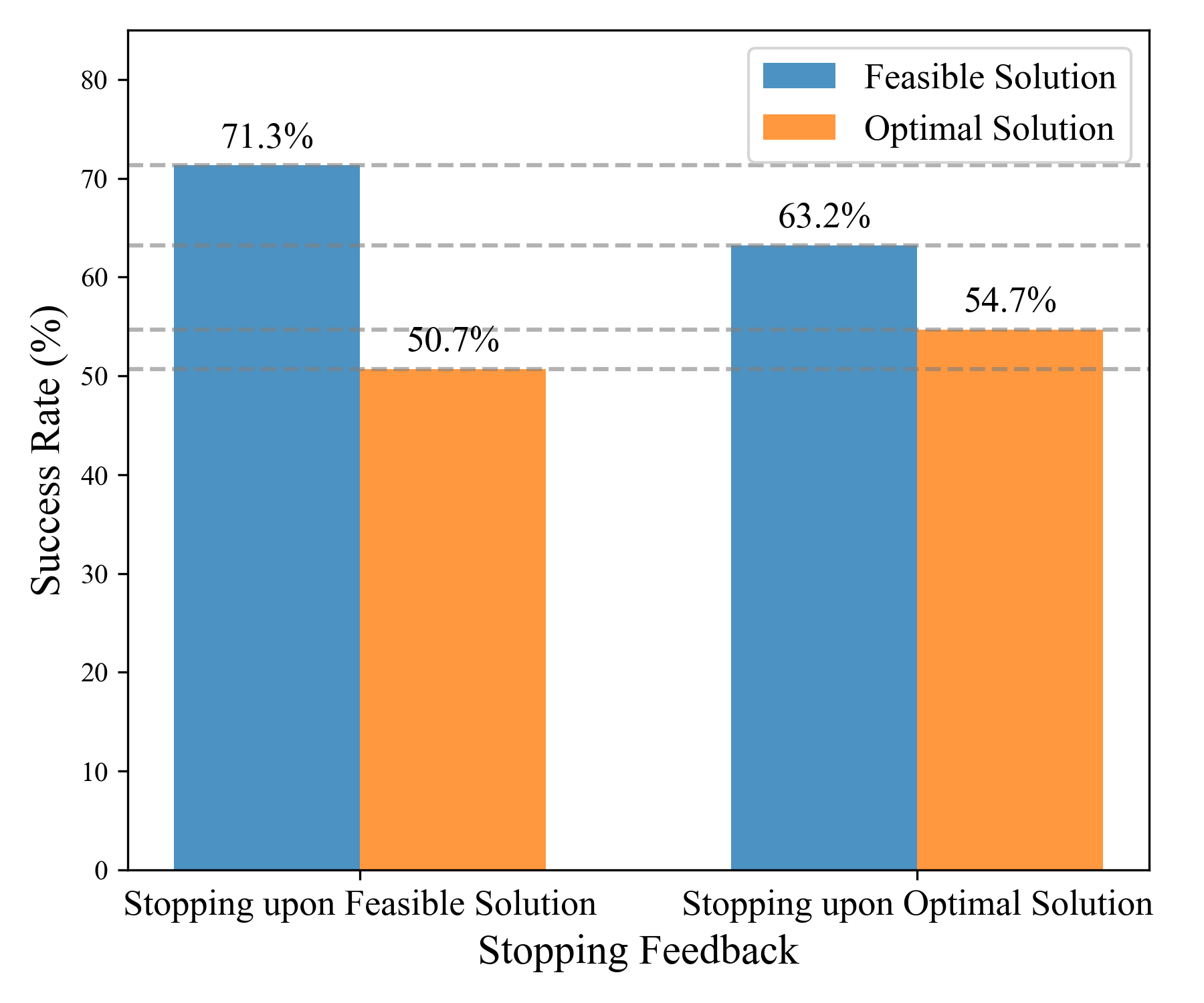}
	\caption{Results across different feedback strategies.}
	\label{fig4}
	
\end{figure}

\subsection{Analysis of Augmented Feedback Strategies}

To address diverse planning requirements, we design multiple feedback strategies tailored to different objectives, including feasibility and optimality. We further introduce two stopping strategies, where termination occurs upon finding either a feasible plan or an optimal plan, and analyze their impact on feasibility and optimality coverage. As shown in Fig. \ref{fig4}, the two stopping strategies exhibit distinct performance trade-offs. Stopping upon finding a feasible solution yields 71.3\% feasibility coverage and 50.7\% optimality coverage. In contrast, stopping only upon reaching an optimal solution reduces feasibility coverage to 63.2\% but improves optimality coverage to 54.7\%. This is due to the iterative refinement in optimality-based stopping, where dynamic plan adjustments may introduce conflicts or violations, rendering some feasible solutions invalid. Meanwhile, the results show that the symbolic verifier provides fine-grained guidance through iterative feedback, promoting the convergence of feasible solutions toward optimality and allowing flexible adaptation to diverse planning requirements.

\section{CONCLUSION}

In this paper, we propose a symbolic feedback-driven iterative self-refinement framework to enhance the robustness and reliability of LLMs in long-horizon planning. A natural language prompting mechanism maps PDDL symbols into textual descriptions, enabling LLMs to better capture task constraints and semantics. We further design a symbolic verifier that translates errors into corrective instructions interpretable by the LLMs, providing precise guidance for iterative self-refinement. In addition, the verifier incorporates a plan recognizer to infer goal reachability and facilitate symbolic knowledge transfer, thereby steering the planning process more effectively toward intended goals. Empirical results demonstrate that the proposed framework substantially improves performance in representative long-horizon planning tasks. Overall, it improves both the feasibility and correctness of LLM-based planning, highlighting its potential to enhance security and support the development of more reliable AI systems. Future work will extend the evaluation of the proposed framework to a broader range of open and dynamic environments.


\section*{Acknowledgment}
This work was supported by the Strategic Priority Research Program of Chinese Academy of Sciences under Grant XDA0480301.


\bibliographystyle{IEEEtran}
\bibliography{reference-final}

\end{document}